\documentclass[10pt,twocolumn,letterpaper]{article}

\usepackage{cvpr}              % To produce the CAMERA-READY version

\usepackage{multirow}
\definecolor{cvprblue}{rgb}{0.21,0.49,0.74}
\usepackage[pdftex, colorlinks=true, linkcolor=blue, citecolor=blue, pdfauthor={Sriram Mandalika}, pdftitle={PRIMEDrive-CoT: A Precognitive Chain-of-Thought Framework for Uncertainty-Aware Object Interaction in Driving Scene Scenario}]{hyperref}
\usepackage[accsupp]{axessibility}

\def\paperID{8} % *** Enter the Paper ID here
\def\confName{CVPR}
\def\confYear{2025}

\title{PRIMEDrive-CoT: A Precognitive Chain-of-Thought Framework for Uncertainty-Aware Object Interaction in Driving Scene Scenario}
\vspace{-7pt}
\author{
Sriram Mandalika \textsuperscript{1}, \quad Lalitha V \textsuperscript{2}, \quad Athira Nambiar \textsuperscript{1*}\\
\textsuperscript{1}
Department of Computational Intelligence,\\
\textsuperscript{2} Department of Electronics and Communication Engineering,\\
Faculty of Engineering and Technology,
SRM Institute of Science and Technology\\
Kattankulathur, Tamil Nadu, 603203, India\\
{\tt\small \{mc9991, lv2876, athiram\}@srmist.edu.in}
}

\begin{document}

\maketitle

\begin{abstract}
Driving scene understanding is a critical real-world problem that involves interpreting and associating various elements of a driving environment, such as vehicles, pedestrians, and traffic signals. Despite advancements in autonomous driving, traditional pipelines rely on deterministic models that fail to capture the probabilistic nature and inherent uncertainty of real-world driving. To address this, we propose \textbf{PRIMEDrive-CoT}, a novel uncertainty-aware model for object interaction and Chain-of-Thought (CoT) reasoning in driving scenarios. In particular, our approach combines LiDAR-based 3D object detection with multi-view RGB references to ensure interpretable and reliable scene understanding. Uncertainty and risk assessment, along with object interactions, are modelled using Bayesian Graph Neural Networks (BGNNs) for probabilistic reasoning under ambiguous conditions. Interpretable decisions are facilitated through CoT reasoning, leveraging object dynamics and contextual cues, while Grad-CAM visualizations highlight attention regions. Extensive evaluations on the DriveCoT dataset demonstrate that PRIMEDrive-CoT outperforms state-of-the-art CoT and risk-aware models.

\end{abstract}

\vspace{-.1cm}
\section{Introduction}

Over recent decades, reasoning architectures such as OpenAI's o1 \cite{openai2023} and DeepSeek R1 \cite{DeepSeekAI2025DeepSeekR1IR} have demonstrated remarkable capabilities in complex decision-making, particularly through techniques like Chain of Thought (CoT) reasoning \cite{wei2022chain}. CoT enables models to break down intricate problems into step-wise reasoning tasks, mimicking structured human cognition. This paradigm has gained significant traction in various application scenarios, such as autonomous driving, robotics and healthcare diagnostics, enhancing interpretability and safety by improving decision-making processes.

Referring to autonomous driving scenarios, the mainstream technical solutions fall under either modular designs or end-to-end driving models\cite{Sachdeva2023Rank2TellAM}. However, these approaches come with a trade-off between system complexity and interpretability. To this end, incorporating the explainability and the reasoning process within end-to-end models was proposed in some recent works \cite{Luo2024PKRDCoTAU, Nie2023Reason2DriveTI, Mandalika2024SegXALEA, Wang_2024_DriveCoT}.  Nevertheless, most of the traditional pipelines rely on deterministic models that fail to capture the inherent probabilistic nature of real-world driving \cite{Zhou2024UATrackUE, Lee2024UncertaintyTrackED}.

We postulate that considering this probabilistic nature is crucial, particularly in high-risk scenarios such as autonomous driving, where precognition — the ability to anticipate and interpret complex situations in advance — plays a pivotal role. Human drivers naturally assess upcoming risks based on contextual observations, inferring potential hazards before they manifest \cite{Tang2022PredictionUncertaintyAwareDF}. Replicating this ability in autonomous systems requires models capable of uncertainty-aware interactions, risk forecasting, and proactive decision-making.

Based on this rationale, we propose a novel CoT-driven object interaction and reasoning framework named PRIMEDrive-CoT, a \textbf{PR}ecognitive \textbf{I}nteraction \textbf{M}odel for \textbf{E}nvironmental Uncertainty in \textbf{Driv}ing Scenarios. Unlike conventional CoT approaches that operate deterministically, PRIMEDrive-CoT incorporates \textit{Bayesian Graph Neural Networks (BGNNs)} to model uncertainty and dynamic object interactions. This provides the system with precognitive capabilities, enabling it to anticipate potential risks and adapt proactively to evolving driving conditions i.e. better handle occlusions, unexpected object behaviours, and complex interactions in dense traffic scenarios. This work brings us closer to realizing \textit{Agentic AI} systems that can perceive, reason, and act autonomously in complex real-world environments. By integrating \textit{uncertainty estimation} and \textit{Chain-of-Thought reasoning}, our end-to-end model effectively balances robustness and interpretability, ensuring safer and more 
reliable decision-making. Furthermore, inspired by SegXAL\cite{Mandalika2024SegXALEA}, an explainable active learning framework for semantic segmentation, our approach leverages human-in-the-loop reasoning to refine predictions in ambiguous cases. This allows for the adaptive incorporation of human expertise into model learning, ensuring that uncertain cases are resolved in a structured, interpretable manner. 

To evaluate the effectiveness of \textbf{PRIMEDrive-CoT}, extensive experiments are carried out on the DriveCoT dataset \cite{Wang_2024_DriveCoT}, specifically targeting scenarios where uncertainty-aware reasoning is critical. The results demonstrate that PRIMEDrive-CoT outperforms existing CoT and uncertainty-driven models, maintaining robustness in challenging conditions like low light and adverse weather, while enhancing situational awareness and real-time adaptability in autonomous driving. The major contributions of this paper are as follows:
\begin{itemize}
    \item Proposal of \textbf{PRIMEDrive-CoT}, a precognitive framework based on object interaction and reasoning based on uncertainty, motivated by human cognition patterns.
    \item Effective decision-making by modelling vehicle-to-pedestrian and vehicle-to-vehicle interactions employing \textbf{Bayesian Graph Neural Networks (BGNNs) utilizing CoT annotations}.
    \item Proposal of an \textbf{proximity-aware risk computation metric} that enables the vehicle to prioritize objects of concern, rather than treating all known objects equally.
\end{itemize}

The rest of the paper is organized as follows: The related works are described in Section 2. The proposed PRIMEDrive-CoT framework is presented in Section 3. The experimental setup and the results are discussed in detail in Section 4 and Section 5 respectively. Finally, the summary of the paper and some future plans are enumerated in Section 6.

\section{Related Works}

\subsection{Chain-of-Thought for Autonomous Reasoning}
Recent advancements in Chain-of-Thought (CoT) reasoning \cite{Wei2022ChainOT} have significantly influenced autonomous driving by enabling vehicles to sequentially decompose complex scenarios, improving decision clarity and interpretability. DriveCoT \cite{Wang_2024_DriveCoT} introduced a dataset specifically designed to train models in generating explicit reasoning traces behind driving decisions, providing step-wise justifications beyond traditional perception pipelines. Similarly, PKRD-CoT \cite{Luo2024PKRDCoTAU} employs a zero-shot prompting approach to integrate CoT reasoning within multi-modal large language models (MLLMs), leveraging pre-trained models and external knowledge bases to enhance context-aware decision-making in autonomous systems beyond sensor-based inputs.

Building on recent advancements, LC-LLM \cite{Peng2024LCLLMEL} is the first approach leveraging Large Language Models (LLMs) for lane-change intention and trajectory prediction in autonomous driving. By framing lane-change prediction as a language modeling task, it integrates Chain-of-Thought (CoT) reasoning to enhance both accuracy and interpretability. Experiments on the highD dataset show substantial improvements in predicting lane-change intentions and trajectories while ensuring transparent explanations.

Expanding LLMs' role in decision-making, RDA-Driver \cite{Huang2024MakingLL} employs multimodal LLMs with reasoning-decision alignment to correct inconsistencies in CoT reasoning and planning. A contrastive loss ensures logical consistency, leading to improved interpretability and reliability, achieving state-of-the-art results on nuScenes and DriveLM-nuScenes. Similarly, Sce2DriveX \cite{Zhao2025Sce2DriveXAG} bridges scene understanding with vehicle control using a cognitive reasoning approach, integrating Bird’s-Eye-View (BEV) maps and local video data for enhanced spatiotemporal perception. Supported by a novel Visual Question Answering (VQA) dataset, it demonstrates superior generalization and top-tier performance on the CARLA Bench2Drive benchmark.

\subsection{Uncertainty-Aware Object Interaction and Risk Assessment in Driving}

In dynamic driving environments, uncertainty-aware models capture occlusion, motion, and interaction-based uncertainties, while learning-based risk assessment enhances decision-making—addressing the limitations of conventional rule-based safety checks that often fail in unstructured scenarios due to unreliable black-box motion predictions \cite{Tang2022PredictionUncertaintyAwareDF}. Uncertainty estimation techniques such as Integrated Gradients \cite{Sundararajan2017AxiomaticAF} and SmoothGrad \cite{Smilkov2017SmoothGradRN} help identify objects most influential to model predictions, improving risk assessment. Additionally, Reason2Drive introduced a large-scale dataset of video-text pairs to train generative models for real-time, interpretable driving explanations \cite{Nie2023Reason2DriveTI}, enhancing transparency in risk perception. Recent approaches, such as Waymo's EMMA model \cite{Hwang2024EMMAEM}, leverage vision-language models for improved decision-making but face high computational costs, limiting real-time deployment. In contrast, Wayve’s camera-only system \cite{Zhong2024EvaluationOO} learns driving behaviour from large-scale videos, improving adaptability but lacking explicit reasoning mechanisms. 

In contrast to the aforementioned approaches, our PRIMEDrive-CoT framework bridges this gap by addressing the lack of uncertainty-aware reasoning in existing CoT-based driving models, which often fail to capture probabilistic interactions and risk factors in dynamic environments. By integrating uncertainty-aware object interaction analysis with explainability-driven risk assessment, our framework ensures robust and interpretable decision-making in complex driving scenarios. Unlike previous methods that rely on deterministic reasoning or black-box neural networks, our approach explicitly models uncertainty leverages Bayesian Graph Neural Networks (BGNNs) for interaction-aware inference and incorporates Grad-CAM visualizations to enhance transparency, setting a new benchmark for risk-aware autonomous decision-making.

\begin{figure*}[t!]
    \centering
    \includegraphics[width=1.05\textwidth]{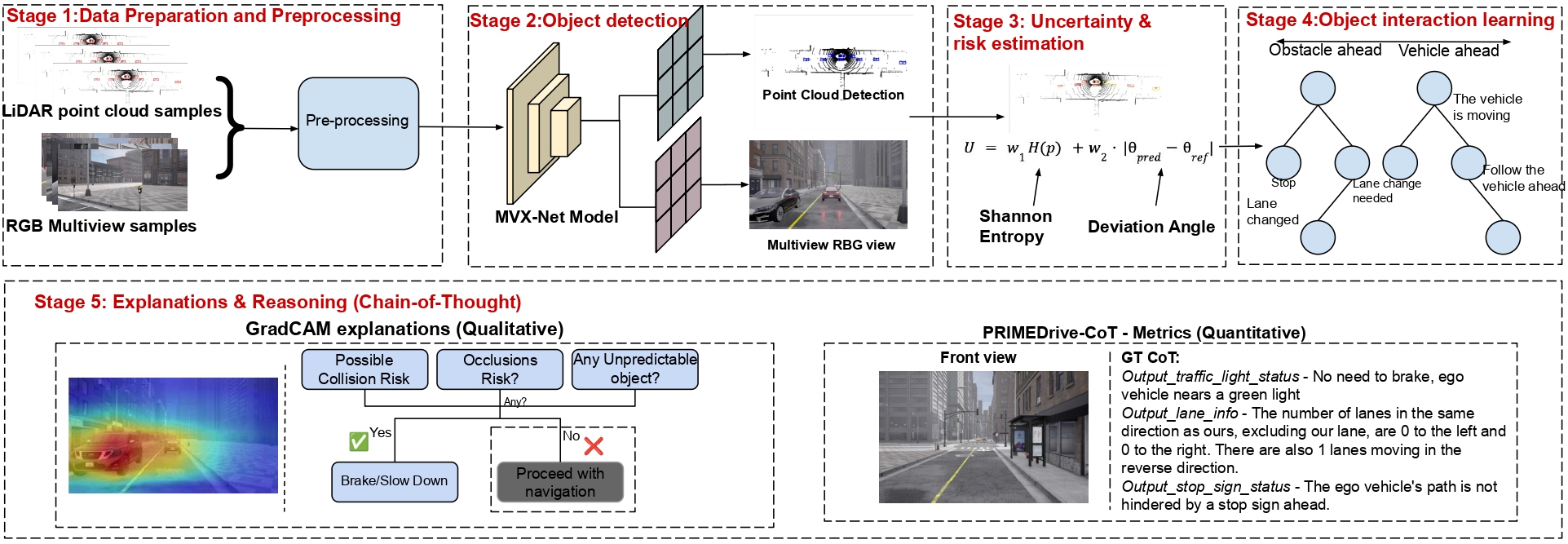}
    \caption{Overview of our proposed \textcolor{black}{\textbf{PRIMEDrive-CoT}} framework. The pipeline consists of Stage 1 (Sec \ref{sec:3.1}):) Data Preprocessing, Stage 2 (Sec \ref{sec:3.2}):) 3D object detection, Stage 3 (Sec \ref{sec:3.3}):) Uncertainty \& Risk assessment, Stage 4 (Sec \ref{sec:3.4}):) Object Interaction Learning and Stage 5 (Sec \ref{sec:3.5}):) Chain-of-Thought (CoT) reasoning and explanation.}
    \label{fig:architecture}
\end{figure*}

\section{Methodology: PRIMEDrive-CoT}

Our proposed PRIMEDrive-CoT framework consists of multiple interconnected components designed to detect, reason, and act in dynamic traffic scenarios, as depicted in Fig.\ref{fig:architecture}. Each of these modules i.e. LiDAR-based 3D object detection, uncertainty estimation, object interaction modelling, and Chain-of-Thought (CoT) reasoning are explained in detail in the forthcoming sections.

\subsection{LiDAR \& Image Processing}
\label{sec:3.1}

The PRIMEDrive-CoT pipeline utilizes two input modalities i.e. LiDAR point clouds and multi-view RGB images. Each of the modalities undergoes systematic preprocessing that lay the foundation for effective sensor fusion, enabling our network to extract complementary features from both modalities while minimizing variability.

\noindent \textbf{LiDAR Preprocessing:}  
Raw LiDAR data, consisting of point measurements $\{(x_i, y_i, z_i, I_i)\}_{i=1}^N$, is first voxelized into a 3D grid. The spatial coordinates are aggregated to form a representative point in each voxel of dimensions $(\Delta x, \Delta y, \Delta z)$. For example, the voxel’s centroid is computed as: 
\vspace{-.3cm}
\[
\bar{x} = \frac{1}{N}\sum_{i=1}^{N} x_i,\quad \bar{y} = \frac{1}{N}\sum_{i=1}^{N} y_i,\quad \bar{z} = \frac{1}{N}\sum_{i=1}^{N} z_i.
\]
Intensity values $I_i$ are normalized to the range $[0, 1]$, typically via a min-max normalization:
\vspace{-.1cm}
\[
I_{\text{norm}} = \frac{I - I_{\min}}{I_{\max} - I_{\min}},
\]
where $I_{\min}$ and $I_{\max}$ are predetermined thresholds based on sensor characteristics. Furthermore, to account for the variable range of LiDAR returns, the coordinates are normalized using a maximum range $R_{\max}$:
\[
x_{\text{norm}} = \frac{x}{R_{\max}},\quad y_{\text{norm}} = \frac{y}{R_{\max}},\quad z_{\text{norm}} = \frac{z}{R_{\max}}.
\]
These steps produce a compact and normalized representation of the 3D scene:  $\{(x_{\text{norm}}, y_{\text{norm}}, z_{\text{norm}}, I_{\text{norm}})\}_{i=1}^N$

which serves as the input for subsequent feature extraction.

\noindent \textbf{Image Preprocessing:}  
Each RGB image captured from our multi-view stereo camera system is resized to a fixed resolution of $224 \times 224$ pixels using \textcolor{black}{bilinear interpolation \cite{Simonyan2014VeryDC}}. To leverage pre-trained deep neural networks (e.g., ResNet), the images are normalized using the ImageNet statistics:
\[
I_{\text{norm}} = \frac{I - \boldsymbol{\mu}}{\boldsymbol{\sigma}},
\]
where $\boldsymbol{\mu} = [0.485,\, 0.456,\, 0.406]$ and $\boldsymbol{\sigma} = [0.229,\, 0.224,\, 0.225]$ represent the mean and standard deviation per color channel, respectively. This normalization ensures consistency across different images and aligns with the training regime of standard CNN backbones.

\subsection{3D Object Detection with LiDAR and Multi-View Images}
\label{sec:3.2}
After the LiDAR and image preprocessing, objects in the driving scene are detected using a multi-modal object detection module. We employ \textcolor{black}{MVX-Net \cite{Sindagi2019MVXNetMV}}, which integrates LiDAR point clouds and multi-view RGB images, to improve object detection robustness, especially in challenging driving environments.

Traditional LiDAR-based detection methods provide accurate 3D spatial localization but lack semantic understanding, making it difficult to differentiate object types in ambiguous conditions. On the other hand, RGB images offer rich texture and colour information but lack precise depth perception, leading to unreliable spatial estimates. By fusing LiDAR and RGB-based features, we leverage the strengths of both modalities, enabling more accurate object localization and identification. Note that, while LiDAR remains the primary detection modality, multi-view RGB images serve as a verification tool, ensuring alignment between detected objects and their visual representations.

Our detection pipeline consists of two main components: a LiDAR backbone based on an enhanced VoxelNet \cite{Zhou2017VoxelNetEL} module and an image backbone utilizing a ResNet34 network pre-trained on ImageNet. The VoxelNet module first voxelizes raw LiDAR point clouds and extracts spatial features, preserving geometric relationships between objects. Simultaneously, the ResNet34 backbone encodes high-level semantic information from the multi-view RGB images. These representations are projected into a common latent space and fused via a multi-layer perceptron (MLP), allowing joint reasoning over both modalities. This fusion mechanism enhances detection robustness in scenarios where individual modalities may be unreliable, such as low-light conditions or occluded objects.

The network predicts the 3D bounding box parameters for each detected object, including its center $(x, y, z)$, dimensions (length $l$, width $w$, height $h$), and yaw angle $\theta$. The regression loss function is defined as:
\vspace{-.3cm}
\begin{equation}
    L_{\text{reg}} = \frac{1}{N} \sum_{i=1}^{N} \| \hat{b}_i - b_i \|_2^2,
    \vspace{-.3cm}
\end{equation}
where $\hat{b}_i$ and $b_i$ are the predicted and ground-truth bounding box parameters, respectively, and $N$ is the total number of objects.

By integrating LiDAR and RGB features, our framework achieves high-precision 3D detection while maintaining uncertainty-awareness, ensuring more interpretable and reliable decision-making for autonomous vehicles.

\subsection{Uncertainty \& Risk assessment}
\label{sec:3.3}
After the 3D object detection stage, the next critical step is evaluating the reliability of the detected objects, as detection confidence may fluctuate due to factors like sensor noise, occlusions, and environmental conditions. While LiDAR provides precise spatial localization, object uncertainties must be accounted for to ensure robust downstream reasoning and decision-making. Fig.~\ref{fig:architecture} (Stage 3) illustrates the uncertainty estimation process integrated into our framework.

\subsubsection{Uncertainty Computation}
To quantify the uncertainty in our detection predictions, we define an uncertainty metric $U$ that accounts for both classification ambiguity and spatial inconsistency. Specifically, we incorporate \textit{Shannon entropy} to measure classification uncertainty and \textit{deviation angle} to assess the inconsistency in predicted object orientation.

\textbf{Shannon entropy} quantifies the confidence of the model’s class predictions by computing the uncertainty in the probability distribution of detected object categories. A higher entropy value indicates greater classification ambiguity. The entropy for a given probability distribution $p$ is defined as:
\vspace{-.3cm}
\begin{equation}
    H(p) = - \sum_{i} p_i \log p_i.
\end{equation}
This captures the degree of uncertainty in classification, ensuring that objects with ambiguous predictions are identified. In addition to classification uncertainty, we introduce \textbf{deviation angle} as a measure of spatial inconsistency. This represents the absolute difference between the predicted yaw angle $\theta_{\text{pred}}$ of an object and a reference orientation $\theta_{\text{ref}}$. A larger deviation suggests greater uncertainty in estimating the object’s orientation. This is formulated as:
\vspace{-.2cm}
\begin{equation}
    |\theta_{\text{pred}} - \theta_{\text{ref}}|.
\end{equation}
% \vspace{-.1cm}
The overall uncertainty metric $U$ is computed as a weighted sum of these two components:
% \vspace{-.1cm}
\begin{equation}
    U = w_1 H(p) + w_2 \cdot |\theta_{\text{pred}} - \theta_{\text{ref}}|,
\end{equation}
where $w_1$ and $w_2$ are weighting coefficients that balance the contributions of classification and spatial uncertainty. A higher value of $U$ indicates greater uncertainty, potentially flagging an object as ambiguous.

\subsubsection{LiDAR-Based Proximity Risk Computation}
To quantitatively assess the risk associated with uncertainty in object detection, we introduce a \textbf{proximity-aware risk computation metric}. This metric enables the system to prioritize objects that pose a higher threat based on their spatial proximity to the ego vehicle. 
The risk score is computed using the LiDAR point cloud, where each detected object is represented by a set of points $\{(x_i, y_i, z_i)\}_{i=1}^N$. The minimum Euclidean distance between the ego vehicle and an object is first determined as:
\begin{equation}
    d_{\text{min}} = \min_{i} \sqrt{x_i^2 + y_i^2 + z_i^2}.
\end{equation}
A lower $d_{\text{min}}$ signifies a closer object, indicating a higher risk level. To effectively model the decay of risk perception with increasing distance, we employ an exponentially decaying function:
\begin{equation}
    R = \exp\left(-\frac{d_{\text{min}}}{\lambda}\right),
\end{equation}
where $\lambda$ is a scaling parameter that controls sensitivity to proximity. Objects that are closer to the ego vehicle receive higher risk scores, while those further away contribute less to immediate decision-making.

For intuitive visualization, detected objects are colour-coded based on their risk scores, as shown in Fig.~\ref{fig:lidar_qualitative_and_Uncertainty}(b). High-risk objects are highlighted in red, moderate-risk in orange, and low-risk in yellow, aligning with the system’s real-time decision-making and ensuring interpretable scene analysis.

\subsection{Object Interaction Learning}
\label{sec:3.4}

To refine detection and decision-making in dynamic driving scenes, we incorporate object interaction learning through Bayesian Graph Neural Networks (BGNNs) \cite{Hasanzadeh2020BayesianGN}.  Bayesian Graph Neural Networks (BGNNs) combine the principles of Graph Neural Networks (GNNs) and Bayesian Inference to model uncertainty in graph-structured data. BGNNs extend conventional GNNs by modelling uncertainty in both node features and edge interactions, making them well-suited for ambiguous and high-risk scenarios. Each object is represented as a node, and interactions are captured via probabilistic edges, enabling structured and uncertainty-aware reasoning.

In our framework, the interaction between two objects \( i \) and \( j \) is modelled using an interaction energy function:
\begin{equation}
    e_{ij} = \lambda_1 D_{ij} + \lambda_2 \Delta V_{ij} + \lambda_3 I_{ij},
\end{equation}
where \( D_{ij} \) is the relative distance, \( \Delta V_{ij} \) is the velocity difference, and \( I_{ij} \) is the contextual interaction intensity. The coefficients \( \lambda_1, \lambda_2, \lambda_3 \) control the influence of each term.

In our setup, BGNNs reason over the graph of detected objects using both spatial-temporal features and their associated uncertainties, which are initially estimated using Shannon entropy over class probabilities. This propagation of uncertainty-aware interactions supports CoT reasoning by enabling relational inferences, such as slowing down for a braking vehicle or yielding to a pedestrian, thus improving both interpretability and driving safety.

\subsection{Explanation and CoT Reasoning}
\label{sec:3.5}

To provide key insights into object interactions and decision-making, our framework incorporates both textual and visual explanations. Specifically, we employ a chain-of-thought (CoT) module to generate concise textual descriptions of detected interactions, while Grad-CAM-based visualizations highlight critical regions influencing model decisions. These explanations ensure transparency in risk assessment and improve interpretability.

The CoT module generates reasoning-based textual descriptions by analyzing interactions and uncertainty factors. For instance, it identifies high-risk scenarios by considering proximity, velocity changes, and occlusions, producing explanations such as "\emph{High risk due to nearby pedestrian and abrupt deceleration}." These insights enhance situational awareness and provide human-readable justifications for the model’s decisions.

To complement textual reasoning, we apply Gradient-weighted Class Activation Mapping (Grad-CAM) \cite{Selvaraju2016GradCAMVE}, which computes attention heatmaps over input images, highlighting key areas that influence decision-making. As shown in Fig.~\ref{fig:gradcam}, detected objects are overlaid with color-coded bounding boxes indicating uncertainty levels, while Grad-CAM heatmaps visualize attention regions in multi-view images. This helps verify whether the model correctly focuses on critical interacting objects when determining speed adjustments and path planning.

Additionally, our framework supports human-in-the-loop interaction using principles from SegXAL \cite{Mandalika2024SegXALEA}, an explainable active learning paradigm for semantic segmentation. This enables users to provide corrective feedback on model-generated explanations, refining both the reasoning and attention mechanisms over time. By incorporating human insight, our approach strengthens interpretability while maintaining adaptability in dynamic environments.

\section{Experimental Setup}
\subsection{Dataset and Evaluation Protocol}
\textcolor{black}{We evaluate \textbf{PRIMEDrive-CoT} on the DriveCoT dataset~\cite{Wang_2024_DriveCoT}, which contains 1,058 CARLA-simulated scenarios with 36,000 labelled samples including multi-view images, LiDAR point clouds, and chain-of-thought annotations. Following the dataset’s protocol, we use 70\% for training, 15\% for validation, and 15\% for testing. Performance is measured using F1-score for speed decisions, path classification accuracy, and standard 3D detection metrics such as IoU, detection accuracy, and deviation angle.
(Table.~\ref{tab:detection_performance})~\cite{Powers2011EvaluationFP, Rezatofighi2019GeneralizedIO}.}

\subsection{Implementation Details}

We use the CARLA simulator to generate diverse driving scenarios, capturing six synchronized 1600×900 RGB camera streams and 32-lane LiDAR data per frame. Each scenario includes detailed metadata such as scenario type, weather, and time of day. A rule-based expert policy controlled the vehicle during data collection, generating Chain-of-Thought (CoT) labels to reflect interpretable decision-making across complex driving contexts.

The proposed PRIMEDrive-CoT framework is implemented in PyTorch and trained on a dual NVIDIA RTX 4090 setup with 128 GB RAM, requiring approximately 4.5 hours for full convergence. For inference, our model achieves an average runtime of 38 ms per frame (\textasciitilde18.7 FPS) on a single RTX 3090, with a total compute cost of ~41.9 GFLOPs and memory usage under 1.2 GB. The CoT reasoning module is highly lightweight (<1.2 GFLOPs) and does not rely on language models, ensuring real-time deployability even under uncertain or novel conditions.
\begin{figure}[t!]
    \centering
    % First row: GT and Prediction side-by-side
    \begin{subfigure}[b]{0.48\textwidth}
        \centering
        \includegraphics[width=\textwidth]{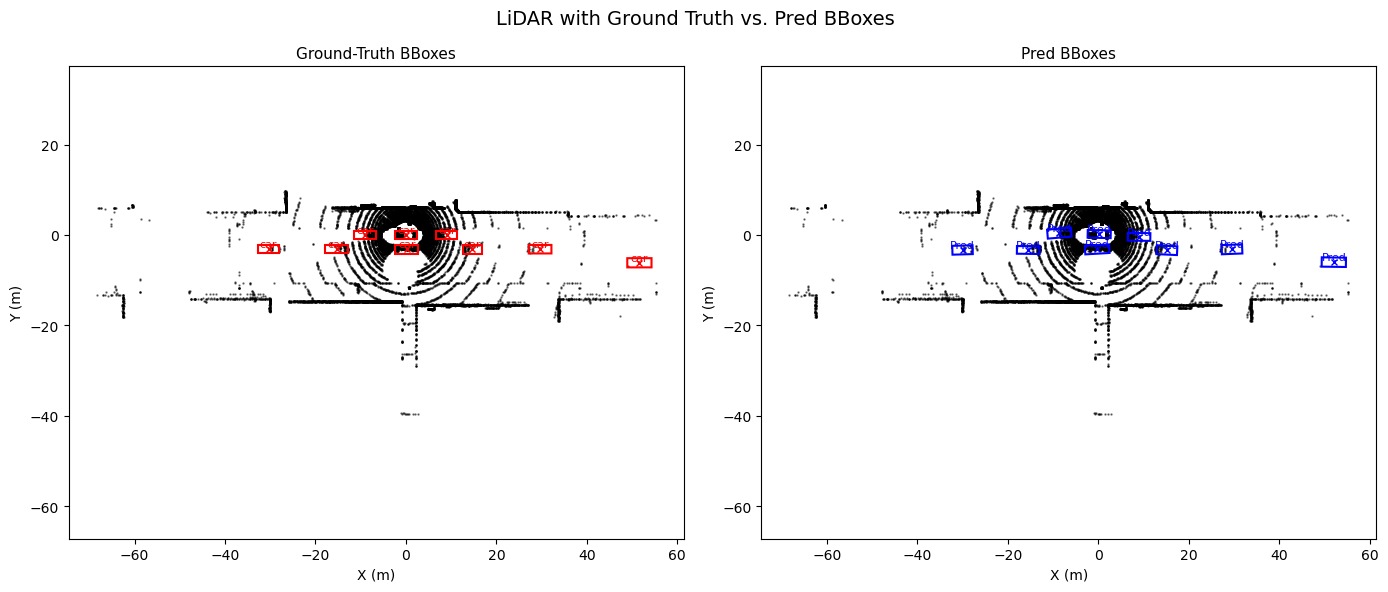}
        \caption{Ground Truth (GT) and Predicted BBs over LiDAR point cloud.}
    \end{subfigure}
    \hfill
    \begin{subfigure}[b]{0.48\textwidth}
        \centering
        \includegraphics[width=\textwidth]{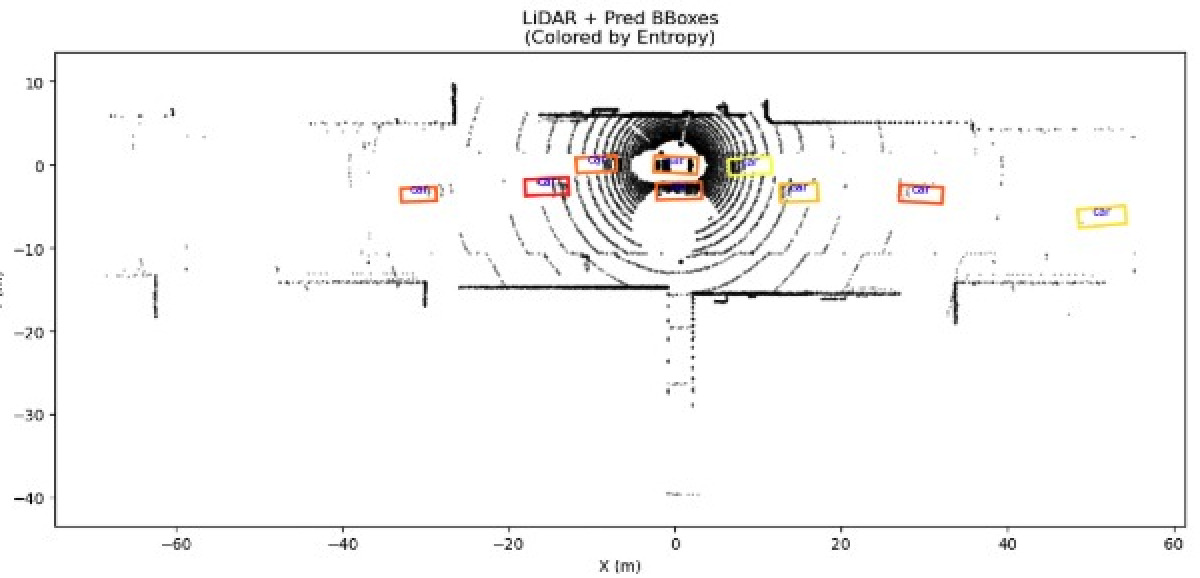}
        \caption{\textcolor{black}{Predicted bounding boxes over LiDAR point cloud with proximity-based uncertainty risk assessment.(Better viewed in colour)}}
    \end{subfigure}
    
    % Common caption for (a) and (b)
    \caption{Qualitative results of LiDAR-based 3D detection and proximity-based uncertainty risk assessment ranking. The predicted bounding boxes (blue) are overlaid on the LiDAR point cloud, while the ground truth (red) serves as a reference.}
    \vspace{-.3cm}
    
    \label{fig:lidar_qualitative_and_Uncertainty}
\end{figure}

\vspace{-.3cm}
\section{Experimental Results}

To verify the effectiveness of our proposed PRIMEDrive-CoT framework, we performed extensive quantitative and qualitative evaluations on the DriveCoT dataset. Table~\ref{tab:comparison} (first row) summarizes our quantitative results, demonstrating that the proposed PRIMEDrive-CoT achieves superior performance in terms of detailed speed decisions (F1-score: 0.85 for SpeedLimit, 0.82 for FollowAhead, 0.79 for SlowDown, 0.78 for SlowApproach, 0.86 for CautiousTurn, and 0.87 for Brake) and waypoint accuracy (87.6\% for Straight, 77.6\% for Turn, and 82.9\% for Lane Change), attributed to the integrated uncertainty-aware reasoning and interaction modelling. These improvements are especially notable in occluded or congested scenes, where interaction-driven reasoning helps mitigate ambiguous predictions.

\begin{table*}[h]
    \centering
    \footnotesize
    \renewcommand{\arraystretch}{1} % Adjust row spacing
    \setlength{\tabcolsep}{2pt} % Adjust column spacing
    \caption{Performance evaluation of PRIMEDrive-CoT on DriveCoT dataset validation split. Previous methods can only extract binary speed decisions (normal drive or brake). Compared to previous methods, the proposed PRIMEDriveCoT can predict more precise and detailed speed decisions and steering waypoints. The PRIMEDriveCoT-Agent outperforms others across multiple categories.}
    \label{tab:comparison}
    \begin{tabular}{l | c c c c c c | c c c}
        \toprule
        \multirow{2}{*}{\textbf{Method}} & \multicolumn{6}{c|}{\textbf{Speed (F1 $\uparrow$)}} & \multicolumn{3}{c}{\textbf{Path (accuracy $\uparrow$ \%)}} \\
        \cmidrule(lr){2-7} \cmidrule(lr){8-10}
         & \small Speed Limit & \small Follow Ahead & \small Slow Down & \small Slow Approach & \small Cautious Turn & \small Brake & \small Straight & \small Turn & \small Lane Change \\
        \midrule
        \textbf{PRIMEDrive-CoT} & 0.85 & \textbf{0.82} & \textbf{0.79} & \textbf{0.78} & \textbf{0.86} & \textbf{0.87} & \textbf{87.6} & \textbf{77.6} & \textbf{82.9} \\
        \midrule
        Transfuser \cite{Chitta2022TransFuserIW} & - & - & - & - & - & 0.10 & 60.6 & 40.1 & 31.1 \\
        TCP \cite{Wu2022TrajectoryguidedCP} & - & - & - & - & - & 0.21 & 63.1 & 42.5 & 29.0 \\
        Interfuser \cite{Shao2022SafetyEnhancedAD} & - & - & - & - & - & 0.35 & 62.6 & 38.1 & 27.3 \\
        direct decision & 0.61 & 0.59 & 0.32 & 0.50 & 0.31 & 0.41 & 84.1 & 74.2 & 75.1 \\
        DriveCoT-Agent\cite{Wang_2024_DriveCoT} & \textbf{0.87} & 0.81 & 0.75 & 0.72 & 0.83 & 0.84 & 87.2 & 76.1 & 79.8 \\
        \bottomrule  % This is after the last row
    \end{tabular}
\end{table*}

\subsection{LiDAR-Based 3D Detection}

To analyse the performance of our LiDAR-based 3D detection, extensive quantitative and qualitative analysis are carried out as shown in Table~\ref{tab:detection_performance} and \textcolor{black}{Fig.~\ref{fig:lidar_qualitative_and_Uncertainty}}. Referring to Table~\ref{tab:detection_performance}, it can be shown that the accuracy of the MVX-Net framework achieved 89.39 \% against the baseline VoxelNet 80.47\%, whereas the previous methods achieved a competitive performance of 87\% \cite{Wang_2024_DriveCoT}. Similarly, the IoU, entropy and F1 score are found to be achieving values such as 78\%, 60\% and 0.76\% respectively. Intuitively, this demonstrates that in our LiDAR-based 3D detection experiments, our enhanced MVX-Net framework accurately localizes objects in complex driving scenarios using an improved VoxelNet architecture for LiDAR feature extraction. The detection process does not rely on RGB input but solely on LiDAR point clouds.

\begin{table}[t!]
    \centering
    \setlength{\tabcolsep}{0.1pt} % Reduce column spacing
    \renewcommand{\arraystretch}{1} % Adjust row spacing
    \caption{Performance of LiDAR-based 3D detection.}
    \label{tab:detection_performance}
    \footnotesize % Use a smaller font size to fit within a single column
    \begin{tabular}{lccccc}
        \toprule
        \textbf{Method} & \textbf{Acc. (\%)} & \textbf{IoU} & \textbf{Entropy ↓} & \textbf{F1 ↑} & \textbf{Dev. Angle (°) ↓} \\
        \midrule
        \textbf{MVX-Net (LiDAR only)}  & 89.39 & 0.78 & 0.42 & 0.85 & 3.7  \\
        \textbf{MVX-Net (LiDAR + RGB)} & 89.39 & 0.78 & 0.42 & 0.85 & 3.7  \\
        \textbf{Baseline VoxelNet}     & 80.47 & 0.67 & 0.60 & 0.76 & 6.1  \\
        \bottomrule
    \end{tabular}
\end{table}

\subsection{Uncertainty Quantification}

The uncertainty of our detection predictions is computed in this section. We compute the Shannon entropy for each detected uncertainty object and the corresponding results as reported in Table \ref{tab:detection_performance}. A higher entropy indicates greater uncertainty, while lower values suggest confident predictions. It can be observed that PRIMEDrive-CoT achieves an overall score of 0.42, compared to the baseline uncertainty value i.e. 0.60. The qualitative analysis of the risk assessment object is colour-coded as red-high risk, orange-moderate risk and yellow-low risk, aligning with the system’s real-time decision-making and ensuring interpretable scene analysis, as shown in Fig.~\ref{fig:lidar_qualitative_and_Uncertainty}(b).

Based on validation analysis, we set a threshold of $0.8$, determined empirically as it consistently aligned with misclassified or low-confidence predictions in the validation set, above which detections are flagged as uncertain for further refinement. This uncertainty metric is critical, as it helps identify challenging or ambiguous scenarios and triggers subsequent refinement stages. By focusing on these high-uncertainty cases, our system can improve its overall detection robustness and ensure downstream decision-making benefits from enhanced confidence measures.

\begin{figure*}[t]
    \centering
    \includegraphics[width=0.85\textwidth]{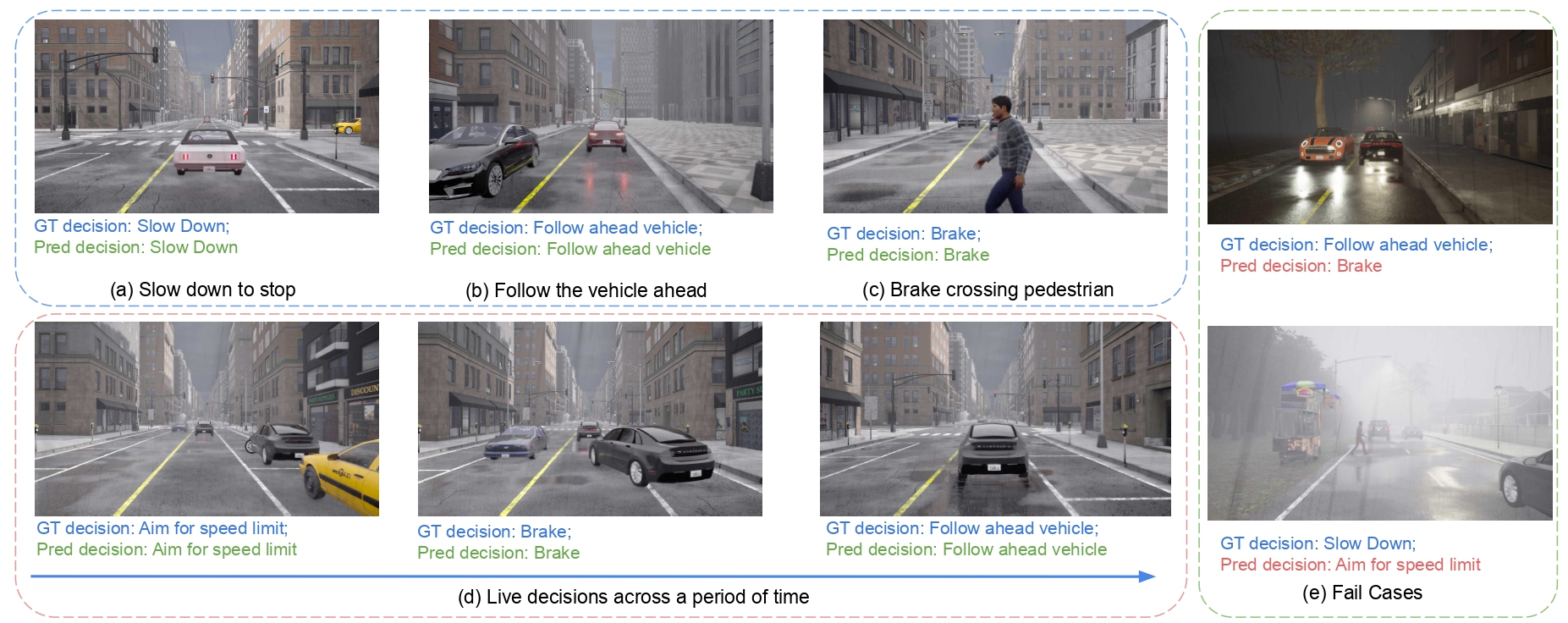}
    \caption{Qualitative results of \textbf{PRIMEDrive-CoT}. The model anticipates and responds to high-risk scenarios, including (a) slowing for static vehicles, (b) following vehicles ahead, (c) braking for pedestrians, and (d) live speed decisions over time. These results demonstrate the role of BGNN-driven interaction reasoning in refining uncertainty and enabling interpretable decisions.}
    \vspace{-.2cm}
    \label{fig:cot_interaction}
\end{figure*}

\begin{figure}[h]
    \centering
    \includegraphics[width=0.45\textwidth]{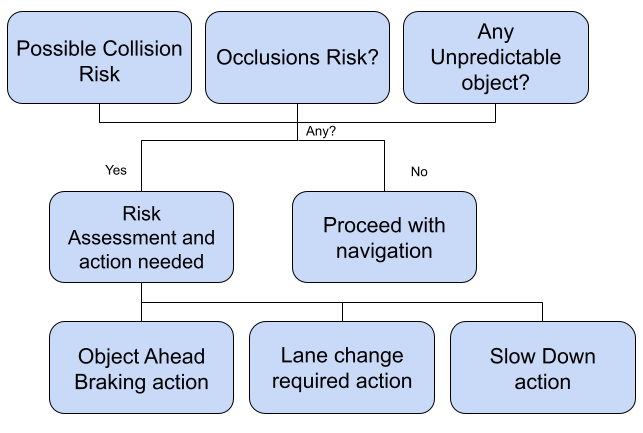}
    \caption{Chain-of-thought (CoT) decision flow corresponding to Fig.~\ref{fig:cot_interaction} for our approach. }
    \label{fig:cot_visual}
\end{figure}

\subsection{Interaction Analysis and Reasoning}
\label{sec:interaction_analysis}

\subsubsection{CoT representations Risk Analysis}
\label{sec:visual_explanations}

\begin{figure}[t!]
    \centering
    % First Row: Original Multi-View Images
    \begin{subfigure}[b]{.5\textwidth}
        \centering
        \includegraphics[width=\textwidth]{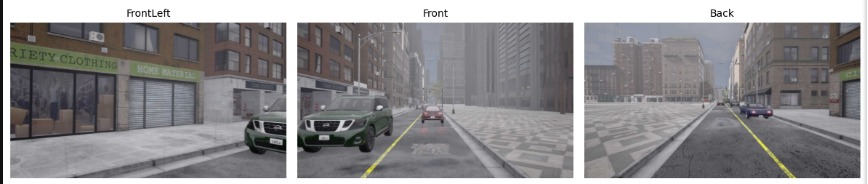}
        \caption{Multi-view camera images from the driving scenario.}
    \end{subfigure}
    
    % Second Row: Grad-CAM Overlays
    \begin{subfigure}[b]{.5\textwidth}
        \centering
        \includegraphics[width=\textwidth]{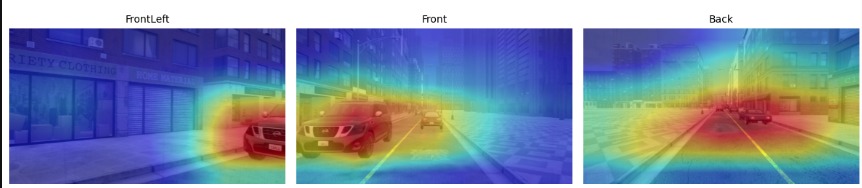}
        \caption{Grad-CAM visualizations highlighting interacting (dynamic) objects, indicating areas of high model attention.}
    \end{subfigure}
    
    \caption{Visualization of interacting objects using Grad-CAM.}
    \label{fig:gradcam}
\end{figure}

Our proposed framework integrates Chain-of-Thought (CoT) reasoning with uncertainty-aware risk assessment for robust decision-making. Fig.~\ref{fig:cot_interaction} and~\ref{fig:cot_visual} illustrate the structured reasoning process and its real-world application. 

\textcolor{black}{Fig.~\ref{fig:cot_interaction} provides qualitative validation, demonstrating the model's adaptive responses to high-risk scenarios. The ego vehicle’s trajectory and predicted waypoints are shown implicitly via lane-following behaviours and alignment with dynamic obstacles. Our BGNN-powered PRIMEDrive-CoT refines uncertainty estimates, ensuring robust, interpretable decision-making across diverse scenarios such as slowing down for traffic, braking for pedestrians, and maintaining safe distances from leading vehicles, as depicted in Fig.~\ref{fig:cot_interaction}.}
% Integration of CoT with Bayesian Graph Neural Networks (BGNNs) refines uncertainty estimates, ensuring robust, interpretable decision-making in dynamic environments. 
Fig.~\ref{fig:cot_visual} outlines the CoT-based decision flow. The system first evaluates key risk factors such as collision risk, occlusions, and unpredictable objects. If any of these factors are present, the framework performs a structured risk assessment and selects the most suitable action: braking for objects ahead, executing a lane change, or slowing down. If no hazards are detected, the ego vehicle proceeds with normal navigation.

\subsubsection{Grad-CAM Explanations}
\label{sec:gradcam_results}

To enhance interpretability, we utilize Grad-CAM to visualize model attention over multi-view RGB images aligned with LiDAR-based detections. This allows human operators to verify predictions in ambiguous scenarios, such as occlusions or noisy sensor returns. As shown in \textcolor{black}{Fig.~\ref{fig:lidar_qualitative_and_Uncertainty}} and Fig.~\ref{fig:gradcam}, RGB overlays provide contextual insight without affecting quantitative performance. Fig.~\ref{fig:gradcam}(a) shows the raw multi-view inputs, while Fig.~\ref{fig:gradcam}(b) highlights Grad-CAM heatmaps, revealing focus on relevant dynamic objects like leading vehicles, pedestrians, and intersection obstacles. Attention shifts correlate with risk-driven decisions e.g., braking for pedestrians or slowing down near occlusions, demonstrating that model outputs are grounded in meaningful interactions. These visual cues validate the reasoning behind decisions and highlight the effectiveness of our uncertainty-aware CoT framework in complex driving environments.

% \vspace{-.4cm}
\subsection{Ablation Study}
\subsubsection{Unverified Multi-View RGB References}
To assess the contribution of multi-view RGB images, we perform an ablation study by removing RGB inputs from the PRIMEDrive-CoT framework. As shown in Table~\ref{tab:detection_performance}, removing RGB has no direct impact on detection accuracy, confirming that RGB is not used for prediction but for verification. However, its absence eliminates Grad-CAM visualizations and cross-modal validation, reducing interpretability—especially in occluded or ambiguous scenes. Thus, while LiDAR ensures strong geometric accuracy, RGB complements it by providing saliency-based justifications for human verification.

\subsubsection{BGNN Hypertuning}
To optimize our Bayesian Graph Neural Network (BGNN), we fine-tune key hyperparameters, including graph depth, embedding dimensions, and adaptive edge weights based on relative velocity and spatial proximity. Results in Table~\ref{tab:bgnn_hypertuning} show that a three-layer BGNN with 128-dimensional embeddings provides optimal performance, improving object classification F1-score by 3.2\% and reducing uncertainty in high-risk detections by 14.5\%. These refinements enhance interaction modelling without additional computational overhead, ensuring robust uncertainty-aware reasoning.

% \subsubsection{Impact of various uncertainty methods}

\begin{table}[t!]
\centering
\caption{Effect of BGNN hyperparameter tuning on detection.}
\label{tab:bgnn_hypertuning}
\footnotesize
\setlength{\tabcolsep}{2pt}
\renewcommand{\arraystretch}{1}
\begin{tabular}{lccc}
\toprule
\textbf{Configuration} & \textbf{F1 Score (\%)} & \textbf{Uncertainty ↓} & \textbf{Notes} \\
\midrule
2-layer, 64-dim & 82.4 & 0.51 & Lower capacity \\
\textbf{3-layer, 128-dim} & \textbf{85.6} & \textbf{0.43} & Optimal configuration \\
4-layer, 256-dim & 85.1 & 0.45 & Higher cost, no gain \\
\bottomrule  % Keep only one \bottomrule
\end{tabular}
\end{table}

\subsection{State-of-the-art Comparison}
Table~\ref{tab:comparison} reports the performance of \textbf{PRIMEDriveCoT-Agent} against state-of-the-art methods on the DriveCoT validation split. Compared with the other end-to-end driving methods such as Transfuser \cite{Chitta2022TransFuserIW} and Interfuser \cite{Wu2022TrajectoryguidedCP},   which require additional supervision (e.g., depth maps or BEV bounding boxes) and produce only binary decisions (normal drive or brake), our framework provides detailed speed decisions via an interpretable chain-of-thought (CoT) reasoning process.

PRIMEDriveCoT-Agent achieves the highest F1 scores across all speed categories, outperforming DriveCoT-Agent~\cite{Wang_2024_DriveCoT} by +4.0\% in slow-down and +3.0\% in cautious turn scenarios, and achieves the best braking accuracy (0.87). For path prediction, our model leads with 87.6\% (straight), 77.6\% (turn), and 82.9\% (lane change) accuracy. These gains could be attributed to the integration of Bayesian Graph Neural Networks (BGNNs) and CoT reasoning, which improve situational awareness and decision robustness under uncertainty.

\section{Conclusion and Future Works}

We introduced PRIMEDrive-CoT, an uncertainty-aware framework that combines BGNNs and CoT reasoning for interpretable and robust autonomous driving. By modeling object interactions, estimating uncertainty with entropy and deviation angles, and using Grad-CAM for visual explanations, our approach delivers strong performance on the DriveCoT benchmark. Extensive evaluations show PRIMEDrive-CoT outperforms existing CoT and risk-aware models, achieving 89\% detection accuracy and improving performance in complex scenarios like occlusions, abrupt braking, and pedestrian crossings. Our method improves slow-down decision F1-score by 4\% and reduces uncertainty by 14.5\% through BGNN-based refinement. 
By integrating structured CoT reasoning with uncertainty modeling, PRIMEDrive-CoT bridges low-level perception and high-level planning, enabling anticipatory, human-aligned driving decisions. This work advances interpretable, risk-aware autonomy in dynamic driving environments. Future work will focus on temporal CoT reasoning, self-supervised learning for better generalization, and optimizing BGNN efficiency for real-time deployment in complex scenarios.

{
    \small
    \bibliographystyle{ieeenat_fullname}
    \bibliography{main}
}


\end{document}